\pdfoutput=1
\documentclass[11pt]{article}
\usepackage[preprint]{acl}
\usepackage{times}
\usepackage{latexsym}
\usepackage[T1]{fontenc}
\usepackage[utf8]{inputenc}
\usepackage{microtype}
\usepackage{graphicx}
\usepackage{booktabs}
\usepackage{amsmath}
\usepackage{amssymb}
\usepackage{caption}

\title{``Who Am I, and Who Else Is Here?''\\Behavioral Differentiation Without Role Assignment\\in Multi-Agent LLM Systems}

\author{Houssam El Kandoussi \\\ Independent Researcher}

\begin{document}
\maketitle

\begin{abstract}
When multiple large language models interact in a shared conversation, do they develop differentiated social roles or converge toward uniform behavior? We present a controlled experimental platform that orchestrates simultaneous multi-agent discussions among 7 heterogeneous LLMs on a unified inference backend, systematically varying group composition, naming conventions, and prompt structure across 12 experimental series (208 runs, 13,786 coded messages). Each message is independently coded on six behavioral flags by two LLM judges from distinct model families (Gemini 3.1 Pro and Claude Sonnet 4.6), achieving mean Cohen's $\kappa = 0.78$ with conservative intersection-based adjudication. Human validation on 609 randomly stratified messages confirmed coding reliability (mean $\kappa = 0.73$ vs.\ Gemini). We find that (1)~heterogeneous groups exhibit significantly richer behavioral differentiation than homogeneous groups (cosine similarity 0.56 vs.\ 0.85; $p < 10^{-5}$, $r = 0.70$); (2)~groups spontaneously exhibit compensatory response patterns when an agent crashes; (3)~revealing real model names significantly increases behavioral convergence (cosine $0.56 \to 0.77$, $p = 0.001$); and (4)~removing all prompt scaffolding converges profiles to homogeneous-level similarity ($p < 0.001$). Critically, these behaviors are absent when agents operate in isolation, confirming that behavioral diversity is a structured, reproducible phenomenon driven by the interaction of architectural heterogeneity, group context, and prompt-level scaffolding.
\end{abstract}

\section{Introduction}

Large language models are increasingly deployed not as isolated responders but as collaborative agents in multi-model architectures \citep{park2023,hong2024,wu2023}. Yet most evaluations focus on individual model capabilities, leaving a critical question unanswered: what happens when heterogeneous LLMs interact as a group?

Human teams exhibit well-documented social phenomena---role differentiation, leadership emergence, compensatory behavior when a member is absent \citep{bales1950,belbin2010}. Recent work on ``machine behavior'' \citep{rahwan2019} suggests that AI systems may develop emergent patterns, but empirical evidence from controlled multi-LLM group interactions remains sparse.

We address these questions through a controlled experimental platform---the War Room---that orchestrates group conversations among 7 LLMs hosted on a unified inference backend (Groq), enabling precise control over model composition, system prompts, naming conventions, and natural agent failure. Across 12 experimental series and 208 completed runs, we collect 13,786 agent messages and code each on six behavioral dimensions using two independent LLM judges from architecturally distinct families, validated against human annotation.

Our five research questions and principal findings are:

\noindent\textbf{RQ1 (Role Differentiation):} All five behavioral trait flags show significant inter-agent variation (Kruskal--Wallis, Bonferroni-corrected $p < 0.05$ for 5/5 flags).

\noindent\textbf{RQ2 (Compensatory Responses):} Groups spontaneously exhibit compensatory response patterns when an agent crashes, with a three-level hierarchy from absence-noting to task redistribution.

\noindent\textbf{RQ3 (Name Bias):} Revealing real model names significantly increases profile convergence (cosine $0.56 \to 0.77$, $p = 0.001$).

\noindent\textbf{RQ4 (Heterogeneity):} Heterogeneous groups show significantly lower profile similarity than homogeneous groups (cosine 0.56 vs.\ 0.85; $p < 10^{-5}$, $r = 0.70$).

\noindent\textbf{RQ5 (Prompt Ablation):} Removing all prompt scaffolding converges profiles to homogeneous-level similarity (K3 cosine $= 0.89$, $p < 0.001$).

Our contributions are: (1)~to our knowledge, the first controlled study specifically isolating behavioral differentiation in heterogeneous multi-LLM groups under minimal prompting; (2)~a reproducible experimental platform with a frozen codebook and dual-judge coding pipeline validated by human annotation; (3)~five empirically grounded findings with effect sizes and confidence intervals; and (4)~methodological innovations including broadcast-tag filtering identified through human validation and multi-family LLM-as-judge panels following \citet{verga2024}.

\section{Related Work}

\paragraph{Multi-Agent LLM Systems.} Recent frameworks orchestrate multiple LLMs for collaborative tasks: MetaGPT assigns software engineering roles \citep{hong2024}, AutoGen enables conversational agents \citep{wu2023}, and ChatDev simulates software companies \citep{qian2024}. These systems prescribe roles through detailed prompts; we study whether roles emerge spontaneously under minimal prompting with no role assignment.

\paragraph{Emergent Behavior in LLM Agents.} \citet{park2023} demonstrated emergent social behaviors in a simulated town of 25 agents using a single model with elaborate memory architectures. \citet{aher2023} used LLMs as proxies for human participants in classic social science experiments. Our work differs by (a)~using heterogeneous model pools rather than a single model, (b)~employing minimal two-line system prompts with no behavioral directives, and (c)~applying rigorous behavioral coding with inter-rater reliability metrics validated against human annotation.

\paragraph{Machine Behavior.} \citet{rahwan2019} proposed studying AI systems through a behavioral science lens. Subsequent work examined LLM personality traits \citep{safdari2025}, social biases \citep{argyle2023}, cooperation in game-theoretic settings \citep{akata2025}, and emergent social conventions in LLM populations \citep{ashery2025}. Ashery et al.\ demonstrated that homogeneous LLM populations spontaneously converge on shared naming conventions; we complement this by studying behavioral \emph{differentiation} in heterogeneous groups under open-ended collaborative tasks.

\paragraph{LLM-as-Judge for Behavioral Coding.} Using LLMs as automated annotators has gained traction \citep{zheng2023,tornberg2023}, with recent work advocating multi-model panels to reduce intra-model bias \citep{verga2024}. We employ a dual-judge panel (Google and Anthropic families) with per-flag Cohen's $\kappa$ \citep{mchugh2012} and conservative intersection-based label adjudication.

\section{Experimental Design}

\subsection{Platform}

The War Room is a browser-based orchestration tool that manages simultaneous API calls to LLMs hosted on Groq's inference platform. Groq provides high-speed inference infrastructure serving unmodified model weights; each model is independently trained by its respective creator. The use of a unified inference backend is a deliberate methodological choice: it ensures identical conditions (latency, API format, error handling) across all agents, eliminating infrastructure-level confounds while preserving the architectural heterogeneity that is the independent variable under study. Model weights are provider-independent: the same LLaMA~3.3~70B checkpoint produces identical outputs whether served by Groq, Together~AI, or a self-hosted instance.

Two fictional project briefs are used: (1)~a food delivery mobile app (\texteuro50,000 budget, 3-month deadline), used in the majority of series; (2)~a cultural open-air festival (\texteuro40,000 budget, 2-month deadline), used in Series~H. The system prompt contains exactly two pieces of information:

\begin{quote}
\small\ttfamily
You are \{nickname\} in a group chat.\\
Other agents: \{list of other nicknames\}
\end{quote}

\noindent The prompt contains \textbf{no} behavioral directives, role assignments, interaction patterns, language requirements, or output format specifications. All series use temperature $= 0.8$ unless otherwise noted (Series~I: 0.95).

\begin{figure}[t]
\centering
\includegraphics[width=\columnwidth]{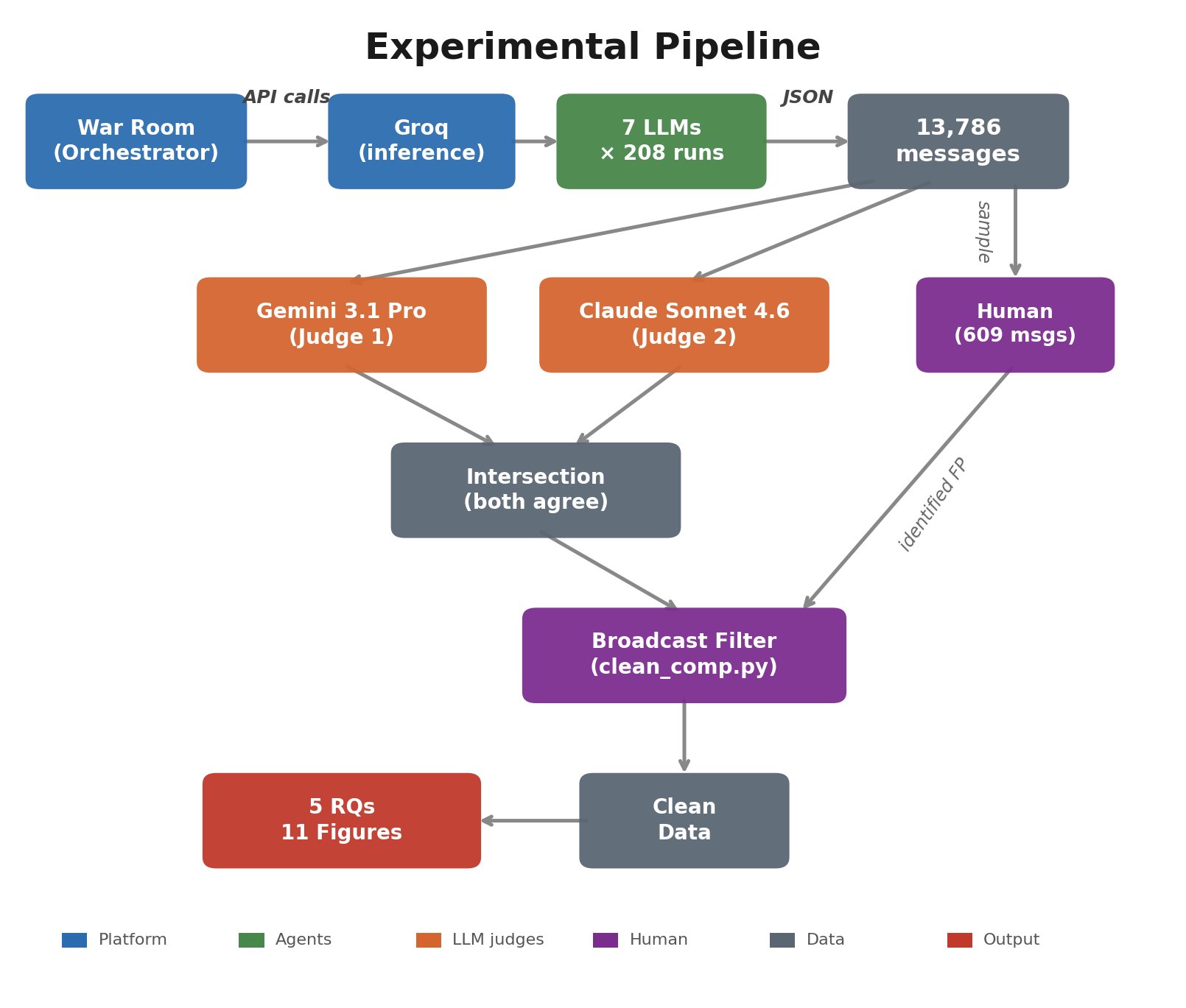}
\caption{Experimental pipeline. The War Room orchestrates 7 LLMs via Groq, producing 13,786 messages across 208 runs. Two architecturally distinct judges (Gemini, Sonnet) independently code each message; conservative intersection yields final labels. Human validation (609 messages) identified broadcast false positives in COMP, corrected by rule-based filtering.}
\label{fig:pipeline}
\end{figure}

\subsection{Agent Pool}

Seven LLMs constitute the heterogeneous pool in Series~A: LLaMA~3.3~70B, GPT-OSS~120B, GPT-OSS~20B, LLaMA~4~Maverick, Kimi~K2, Qwen~3~32B, and LLaMA~4~Scout. An eighth model (DeepSeek~R1, \texttt{deepseek-r1-distill-llama-70b}), deprecated on the Groq platform at the time of data collection, was deliberately included as a controlled failure stimulus: each API call returns a decommissioning notice that is broadcast to the group, enabling observation of compensatory responses (RQ2) under ecologically valid conditions---the failure is genuine, not simulated, but its occurrence is by design.

\subsection{Experimental Series}

\begin{table}[t]
\centering
\small
\begin{tabular}{@{}llrl@{}}
\toprule
\textbf{Ser.} & \textbf{Manipulation} & \textbf{Runs} & \textbf{RQ} \\
\midrule
A & Baseline (heterogeneous) & 21 & 1,2,4 \\
B & Homogeneous (8$\times$ LLaMA 70B) & 21 & 4 \\
C & Real model names & 21 & 3 \\
E & Agent order shuffled & 21 & rob. \\
F & Isolated (no group) & 21 & ctrl. \\
G & English-language brief & 21 & lang. \\
H & Festival task brief & 15 & rob. \\
K1 & No peer list in prompt & 15 & 5 \\
K2 & No agent identifier & 15 & 5 \\
K3 & Empty system prompt & 15 & 5 \\
I & High temperature (0.95) & 11 & par. \\
J & 20 rounds (extended) & 11 & temp. \\
\bottomrule
\end{tabular}
\caption{Experimental series. 208 total runs across 12 conditions. Series~B uses 8 instances of LLaMA~3.3~70B, selected as a mid-range model (neither the largest nor smallest in the pool) to provide a neutral homogeneity baseline.}
\label{tab:series}
\end{table}

\subsection{HARKing Prevention}

Approximately 35 pilot runs were conducted to develop the behavioral taxonomy and calibrate the codebook, which was frozen before confirmatory data collection. No pilot data is included in any reported statistic \citep{kerr1998}.

\subsection{Behavioral Coding}

Each message is coded on five behavioral trait flags (PHATIC, META, LEAD, ARCH, AGREE) plus one reactive flag (COMP) and a structural flag (has\_xref). COMP measures response to an external event (agent failure) and is analyzed separately in RQ2; behavioral variance (RQ4, RQ5) is computed over the five trait flags only.

\subsection{Dual-Judge Coding Panel}

Gemini 3.1 Pro (Google) and Claude Sonnet 4.6 (Anthropic) independently code all 13,786 messages. Final labels use conservative intersection.

\begin{table}[t]
\centering
\small
\begin{tabular}{@{}lrl@{}}
\toprule
\textbf{Flag} & $\kappa$ & \textbf{Level} \\
\midrule
PHATIC & 0.938 & Excellent \\
COMP & 0.856 & Excellent \\
META & 0.773 & Substantial \\
ARCH & 0.730 & Substantial \\
LEAD & 0.712 & Substantial \\
AGREE & 0.700 & Substantial \\
\midrule
\textbf{Mean (6 flags)} & \textbf{0.785} & \textbf{Substantial} \\
\bottomrule
\end{tabular}
\caption{Inter-rater reliability (Cohen's $\kappa$) on 13,786 messages (clean data).}
\label{tab:irr}
\end{table}

\subsection{Human Validation}

A human annotator independently coded 609 messages randomly sampled with proportional stratification across all 12 series and all agent types. Of these, 565 contained at least one positive flag from any rater. This validation identified systematic false positives in COMP coding: broadcast @mentions (e.g., ``@Agent-A, @Agent-B, \ldots\ @Agent-H'') were misclassified as compensation events. A rule-based post-processing filter was applied to all three judge CSVs, removing broadcast tags while preserving genuine compensation references (DeepSeek-specific mentions, absence keywords). After correction, mean $\kappa$ between human and Gemini reached 0.73 (substantial; per-flag values in Appendix~\ref{app:human}).

\section{Results}

All tests use Bonferroni correction. Effect sizes: $r = Z/\sqrt{N}$. Bootstrap CIs: 10,000 resamples (seed $= 42$). Behavioral variance is computed over five trait flags (PHATIC, META, LEAD, ARCH, AGREE); COMP is analyzed separately as a reactive flag.

\subsection{RQ1: Role Differentiation}

\emph{Do agents develop distinct behavioral profiles?}

\begin{figure}[t]
\centering
\includegraphics[width=\columnwidth]{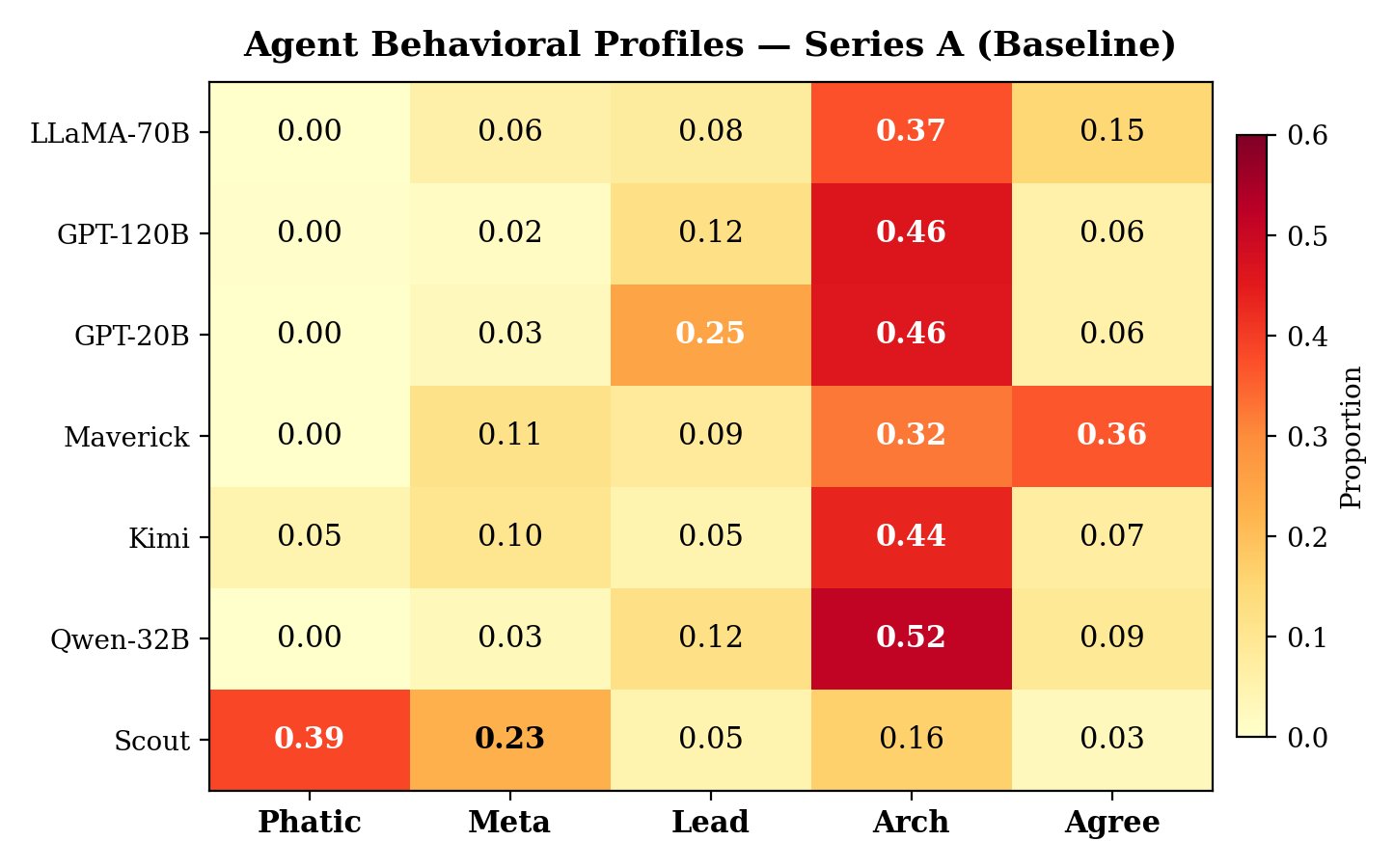}
\caption{Behavioral profiles in Series~A (5 trait flags). Agents spontaneously develop distinct signatures under a two-line prompt with no role assignment.}
\label{fig:profiles}
\end{figure}

\begin{table}[t]
\centering
\small
\begin{tabular}{@{}lrrl@{}}
\toprule
\textbf{Flag} & \textbf{H} & $p_{\text{Bonf}}$ & \textbf{Sig.} \\
\midrule
PHATIC & 46.60 & $< 0.001$ & *** \\
AGREE & 44.52 & $< 0.001$ & *** \\
LEAD & 28.97 & $< 0.001$ & *** \\
ARCH & 24.47 & 0.002 & ** \\
META & 18.13 & 0.030 & * \\
\bottomrule
\end{tabular}
\caption{RQ1: Kruskal--Wallis tests ($\alpha_{\text{Bonf}} = 0.05/5$).}
\label{tab:rq1}
\end{table}

Mean pairwise cosine similarity: $\bar{\cos}\theta_A = 0.56$, indicating markedly differentiated profiles.

\subsection{RQ2: Compensatory Response Patterns After Agent Failure}

\emph{Do groups exhibit compensatory responses when an agent crashes?}

DeepSeek~R1, deliberately included as a controlled failure stimulus (\S3.2), produced systematic API errors across all runs. After filtering broadcast @mentions (validated by human coding), we identify genuine compensation events where agents explicitly reference the crashed model, note its absence, take over its responsibilities, or redistribute its tasks.

\begin{figure}[t]
\centering
\includegraphics[width=\columnwidth]{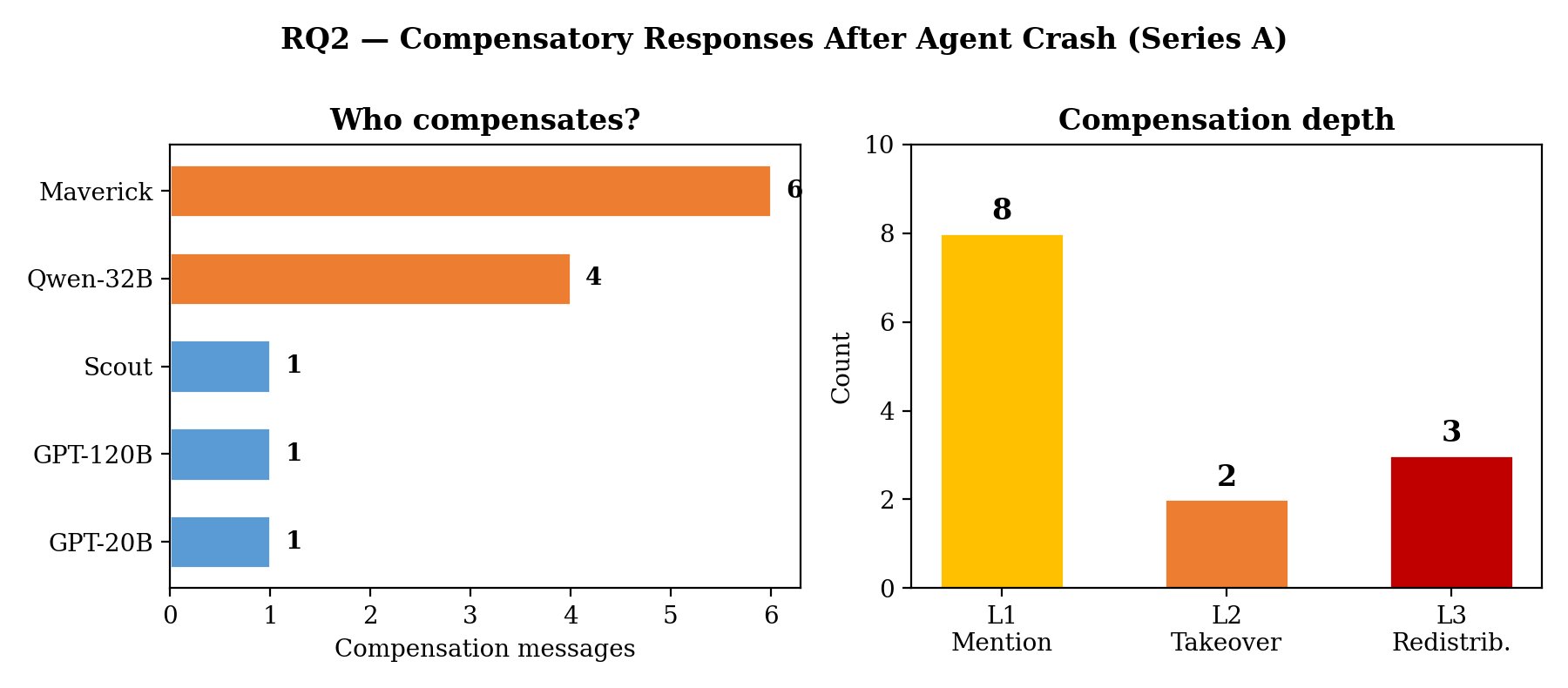}
\caption{Compensation distribution by agent (left) and depth level (right) in Series~A. All events reference DeepSeek specifically.}
\label{fig:comp}
\end{figure}

After conservative filtering validated by human annotation, 166 genuine compensation events are identified across 9 of 12 experimental series, with consistent absence in three conditions: B (no crashed agent in pool), F (no group context), and K3 (no peer awareness in prompt). In the baseline (Series~A), 13 events occur across 6/21 runs, with a three-level hierarchy: L1 (noting absence, 8), L2 (task takeover, 2), L3 (explicit redistribution, 3). Series~C (real names) shows the highest rate at 2.4/run, as agents reference ``DeepSeek R1'' by name. Series~E (shuffle) reaches 3.5/run.

\subsection{RQ3: Name Bias}

\emph{Do naming conventions affect behavioral profiles?}

\begin{figure}[t]
\centering
\includegraphics[width=\columnwidth]{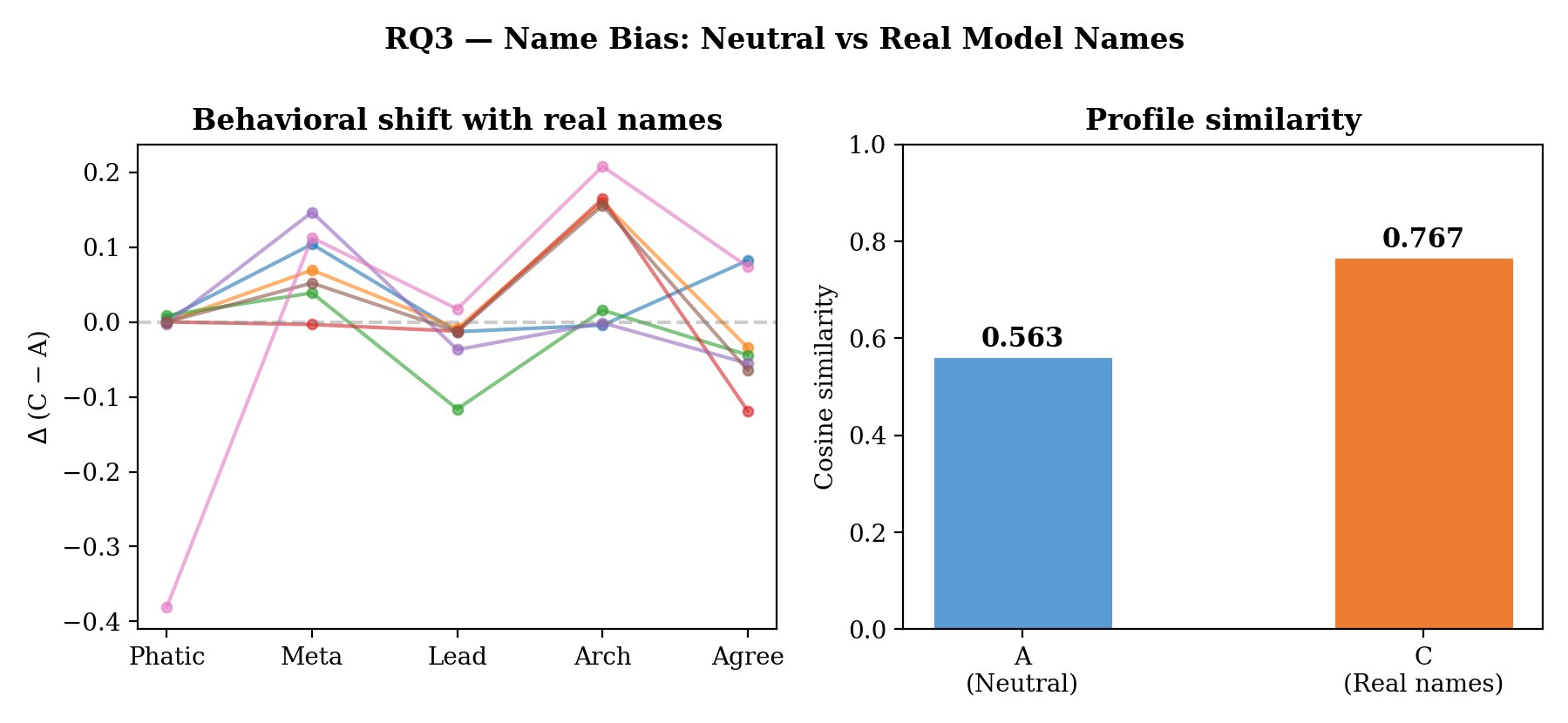}
\caption{Name bias: neutral identifiers (A) vs.\ real model names (C). Higher cosine $=$ less differentiation.}
\label{fig:name}
\end{figure}

Pairwise cosine similarity increases from 0.56 (neutral) to 0.77 (real names), $p = 0.001$, $r = 0.50$. Real model names significantly \emph{reduce} behavioral differentiation.

\subsection{RQ4: Heterogeneity vs.\ Homogeneity}

\emph{Do diverse groups behave differently from homogeneous groups?}

\begin{figure}[t]
\centering
\includegraphics[width=\columnwidth]{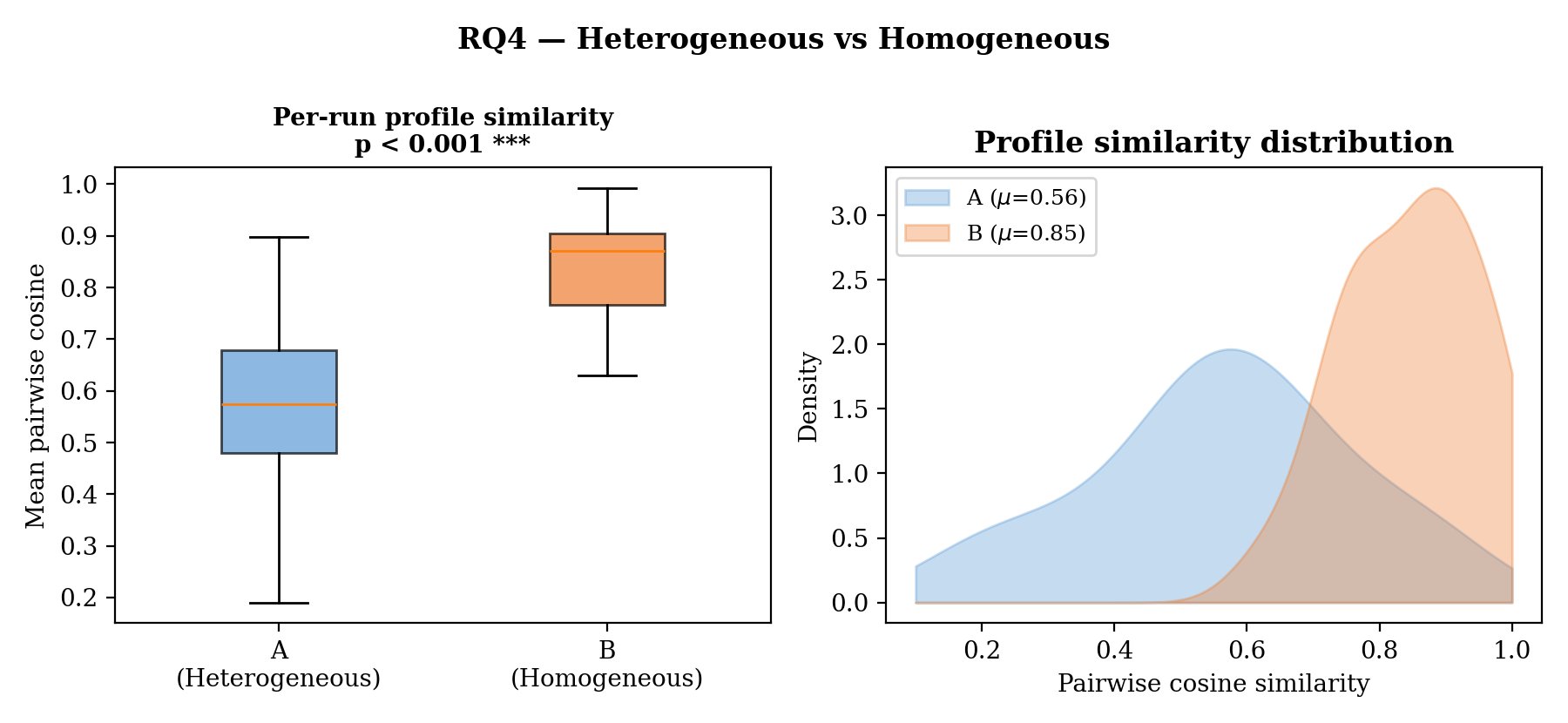}
\caption{Profile similarity: heterogeneous (A) vs.\ homogeneous (B). Non-overlapping bootstrap CIs.}
\label{fig:het}
\end{figure}

\begin{table}[t]
\centering
\small
\begin{tabular}{@{}lccccl@{}}
\toprule
\textbf{Comp.} & \textbf{cos\,A} & \textbf{cos\,X} & $p$ & $r$ & \\
\midrule
A\,vs.\,B & 0.56 & 0.85 & $6{\times}10^{-6}$ & 0.70 & *** \\
A\,vs.\,K3 & 0.56 & 0.89 & ${<}\,0.001$ & 0.68 & *** \\
A\,vs.\,K2 & 0.56 & 0.68 & 0.395 & 0.25 & ns \\
A\,vs.\,K1 & 0.56 & 0.56 & 1.000 & 0.03 & ns \\
\bottomrule
\end{tabular}
\caption{RQ4--RQ5 results on cosine similarity (5 trait flags).}
\label{tab:rq45}
\end{table}

Cosine A $= 0.56$ (CI: [0.48, 0.65]) vs.\ B $= 0.85$ (CI: [0.80, 0.89])---non-overlapping. This is our strongest result.

\subsection{RQ5: Prompt Ablation}

\begin{figure}[t]
\centering
\includegraphics[width=\columnwidth]{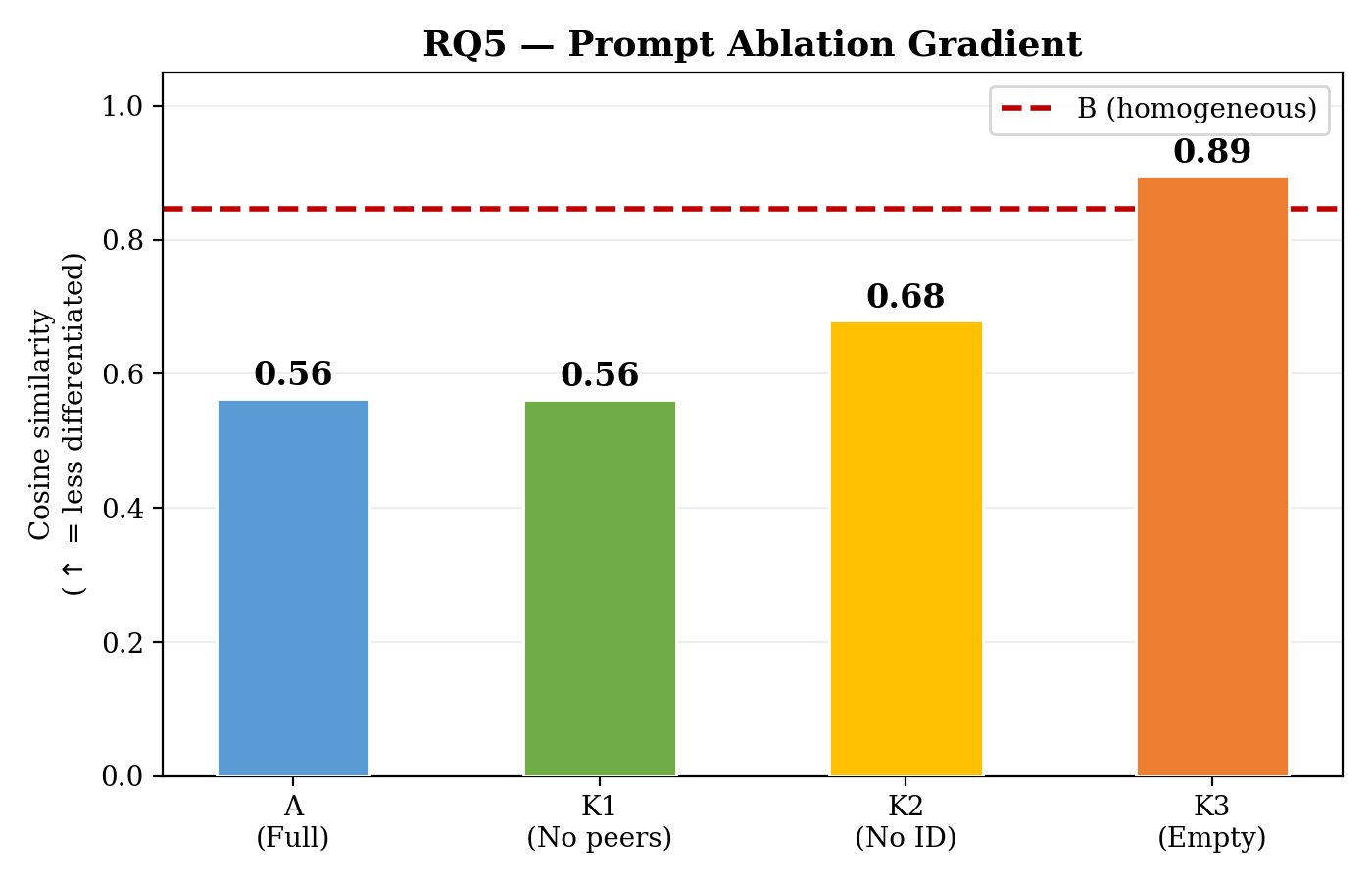}
\caption{Prompt ablation gradient. K3 (empty prompt) converges to homogeneous-level similarity.}
\label{fig:ablation}
\end{figure}

Minimal social scaffolding---a two-line prompt containing only agent identity and peer awareness, with no behavioral directives---is sufficient to activate the full differentiation potential of a heterogeneous group. Removing this context (K3: empty prompt) collapses profiles to homogeneous-level similarity (cosine $= 0.89$, $p < 0.001$, $r = 0.68$). Behavioral diversity thus requires both architectural heterogeneity \emph{and} social scaffolding; together they produce robust differentiation that neither achieves alone.

\begin{center}
\emph{\textbf{Who am I, and who else is here.}}
\end{center}

\subsection{Sensitivity Analysis}

All statistics are recomputed under union labels (Appendix~\ref{app:sensitivity}). RQ4 remains highly significant ($p < 10^{-5}$). RQ3 remains significant ($p < 0.01$). All main conclusions are robust across both aggregation strategies. Additionally, all RQ1 and RQ4 analyses were recomputed excluding Kimi~K2, which exhibited variable participation (8.2\% of Series~A messages vs.\ ${\sim}$15\% expected due to intermittent API errors); all results remain significant with comparable effect sizes (Appendix~\ref{app:kimi}).

\section{Discussion}

\paragraph{Behavioral Differentiation Is Structured.} The significant gap between heterogeneous and homogeneous groups (RQ4), combined with the ablation gradient (RQ5), establishes that behavioral diversity emerges from the interaction between architectural heterogeneity and prompt-level social scaffolding.

\begin{samepage}
\paragraph{Isolation Confirms Emergence.} Series~F provides a critical control: when agents operate in isolation (one agent per run, no group context), behavioral flags are virtually absent---only 4.2\% of messages exhibit any trait, compared to 64.3\% in the group setting (Series~A). Four of five trait flags (PHATIC, META, LEAD, AGREE) drop to exactly zero in isolation. This demonstrates that the behavioral profiles observed in group settings are not pre-existing model defaults but emerge specifically from inter-agent interaction. Even ARCH, the most content-driven flag, drops from 39\% in groups to 4\% in isolation, confirming that the collaborative context elicits technical specificity that models do not produce when operating alone.
\end{samepage}

\begin{figure}[t]
\centering
\includegraphics[width=\columnwidth]{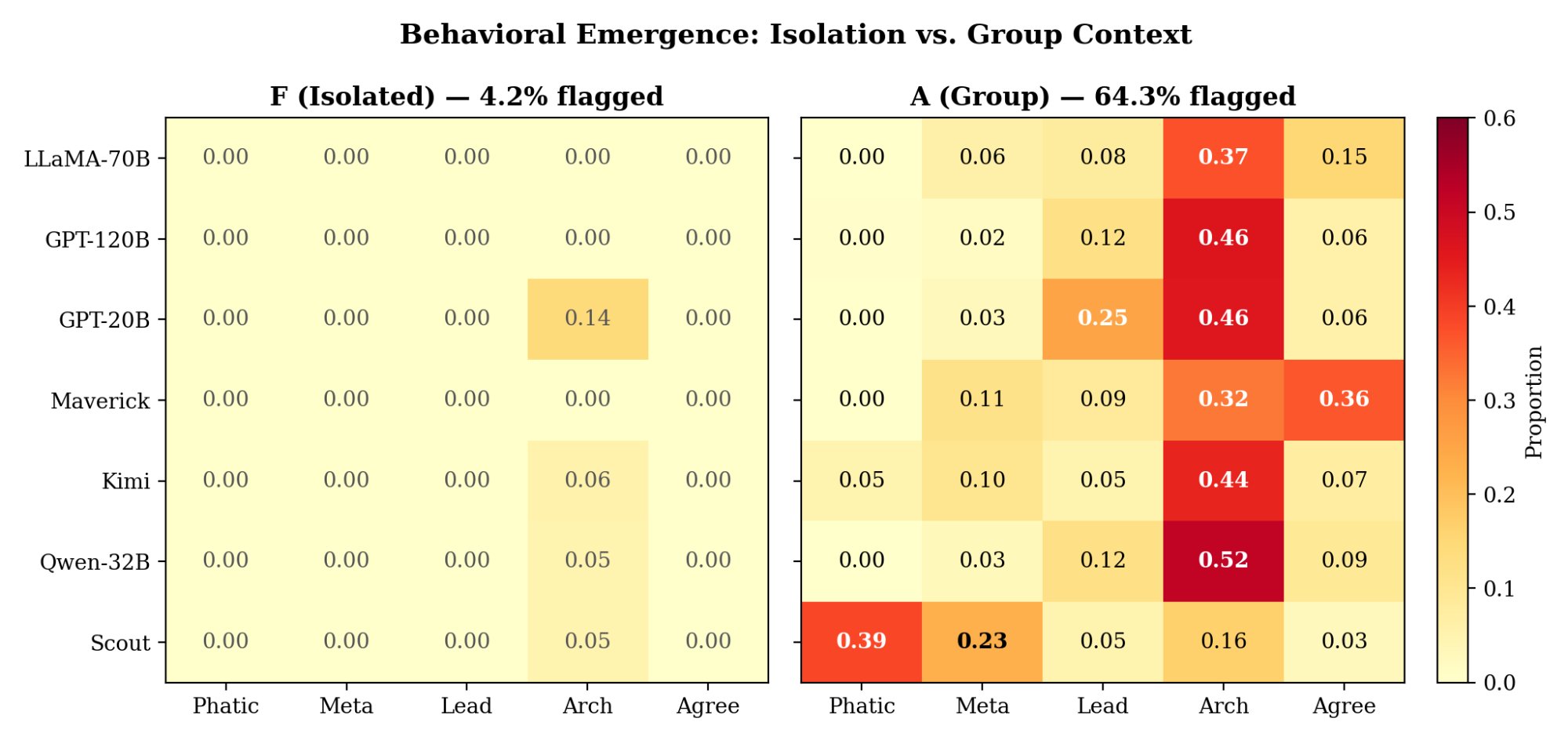}
\caption{Behavioral profiles for the same 7 agents in isolation (Series~F, left) vs.\ group setting (Series~A, right). Social behaviors (PHATIC, META, LEAD, AGREE) are absent in isolation, confirming that differentiation emerges from group interaction.}
\label{fig:isolation}
\end{figure}

\paragraph{Compensatory Responses as Emergent Phenomenon.} The three-level compensatory response pattern (RQ2) emerges without any prompt instruction. Human validation was critical in distinguishing genuine compensation from broadcast @mentions---a methodological contribution that strengthens the finding: the remaining events are exclusively DeepSeek-specific references demonstrating real group awareness.

\paragraph{Naming Conventions as Behavioral Moderators.} When agents see neutral identifiers (Agent-A, Agent-B), they develop distinct behavioral profiles. When the same agents see their real model names (LLaMA~4~Maverick, GPT-OSS~120B), their profiles converge significantly (cosine $0.56 \to 0.77$, $p = 0.001$). Real model names carry identity signals that reduce behavioral individuation: when the system prompt contains real model identifiers instead of neutral labels, all profiles shift toward a uniform technical pattern---ARCH increases across all agents while social differentiation behaviors (LEAD, AGREE) decrease. The convergence is not hierarchical; no single model gains dominance. This convergence effect echoes findings by \citet{baltaji2024}, who observed persona collapse toward a generic mean in multi-agent LLM discussions, albeit in a different setting (assigned cultural personas, single model). Real model names additionally carry compound identity signals---brand association, architecture family, and parameter counts (e.g., `120B' vs.\ `20B')---whose individual contributions cannot be disentangled in the current design, opening a direct line of investigation for future work. The practical implication is immediate: in multi-agent deployments, neutral identifiers preserve the behavioral diversity that heterogeneous pools are designed to provide.

\paragraph{Design Implications for Multi-Agent Systems.} For practitioners designing multi-agent LLM deployments, our findings yield four actionable guidelines: (1)~\emph{Maximize architectural diversity}---heterogeneous pools produce richer behavioral repertoires than homogeneous ones. (2)~\emph{Provide minimal identity scaffolding}---a two-line prompt with agent identity and peer awareness activates full differentiation. (3)~\emph{Use neutral naming conventions}---exposing real model names reduces the diversity that heterogeneity is designed to provide. (4)~\emph{Anticipate compensatory dynamics}---groups naturally redistribute responsibilities when an agent fails, offering a degree of built-in fault tolerance.

\section{Conclusion}

Across 208 runs, 12 series, and 13,786 coded messages, we demonstrate that heterogeneous LLM groups develop significantly richer behavioral differentiation than homogeneous groups, spontaneously exhibit compensatory responses to agent failures, and are shaped by naming conventions and prompt structure. Agents operating in isolation show virtually none of these behaviors, confirming that differentiation is a product of group interaction. Human validation of 609 messages identified and corrected systematic coding errors, strengthening all reported findings.

These results carry direct implications for the design of multi-agent systems. As LLM-based agent architectures scale from single-model pipelines to heterogeneous ensembles, our findings suggest that behavioral diversity is not a byproduct to be suppressed but a resource to be harnessed---through deliberate architectural mixing, minimal identity scaffolding, and neutral naming conventions. The finding that a two-line prompt unlocks the full differentiation potential of a diverse group challenges the assumption that complex role descriptions are necessary for functional specialization in multi-agent systems. More broadly, this work opens a new empirical frontier: the systematic study of machine social behavior under controlled conditions, bridging multi-agent systems research with behavioral science methodology.

\section*{Limitations}

Our findings are bound to the specific model versions available at the time of data collection, and behavioral profiles are expected to evolve as providers update their models. Kimi~K2 exhibited variable participation due to intermittent API errors, producing between 1 and 10 messages per run rather than the expected 11; we retain these runs to preserve ecological validity, as variable agent reliability is inherent to real-world multi-model deployments. This study examines behavioral patterns, not task performance; evaluating the quality of collaborative outcomes (e.g., correctness of proposed architectures or budget estimates) is left for future work. Results may vary with other model pools, task domains, or future model updates; generalization beyond the seven model families tested here should not be assumed without replication.

\section*{Ethics Statement}

No human participants were studied. All interactions analyzed are between commercial LLM APIs processing fictional project briefs. Total API costs: approximately \$300 USD (\$100 Groq inference, \$200 coding judges).

\section*{Data Availability}

All experimental data (208 conversation transcripts in JSON format), the coding pipeline outputs (three judge CSVs and the intersection CSV), the figure generation script, and the War Room orchestration tool are available at \url{https://github.com/elkandoussihoussam/WarRoom}.

\bibliography{references}

\appendix

\section{Codebook}
\label{app:codebook}

\begin{samepage}
Frozen after ${\sim}$35 pilot runs.

\begin{table}[h!]
\centering
\small
\begin{tabular}{@{}lp{5cm}@{}}
\toprule
\textbf{Flag} & \textbf{Definition} \\
\midrule
PHATIC & Under 15 words, no substantive content. \\
META & Comments on the conversation process. \\
LEAD & Assigns a task to a named agent (not self). \\
ARCH & Names 2+ specific technologies/tools. \\
AGREE & Explicit agent name + agreement expression. \\
COMP & References the crashed agent specifically (not broadcast @mentions). L1\,=\,notes absence; L2\,=\,covers work; L3\,=\,redistributes. \\
\bottomrule
\end{tabular}
\end{table}
\end{samepage}

\section{Confusion Matrices}
\label{app:confusion}

\noindent\includegraphics[width=0.92\columnwidth]{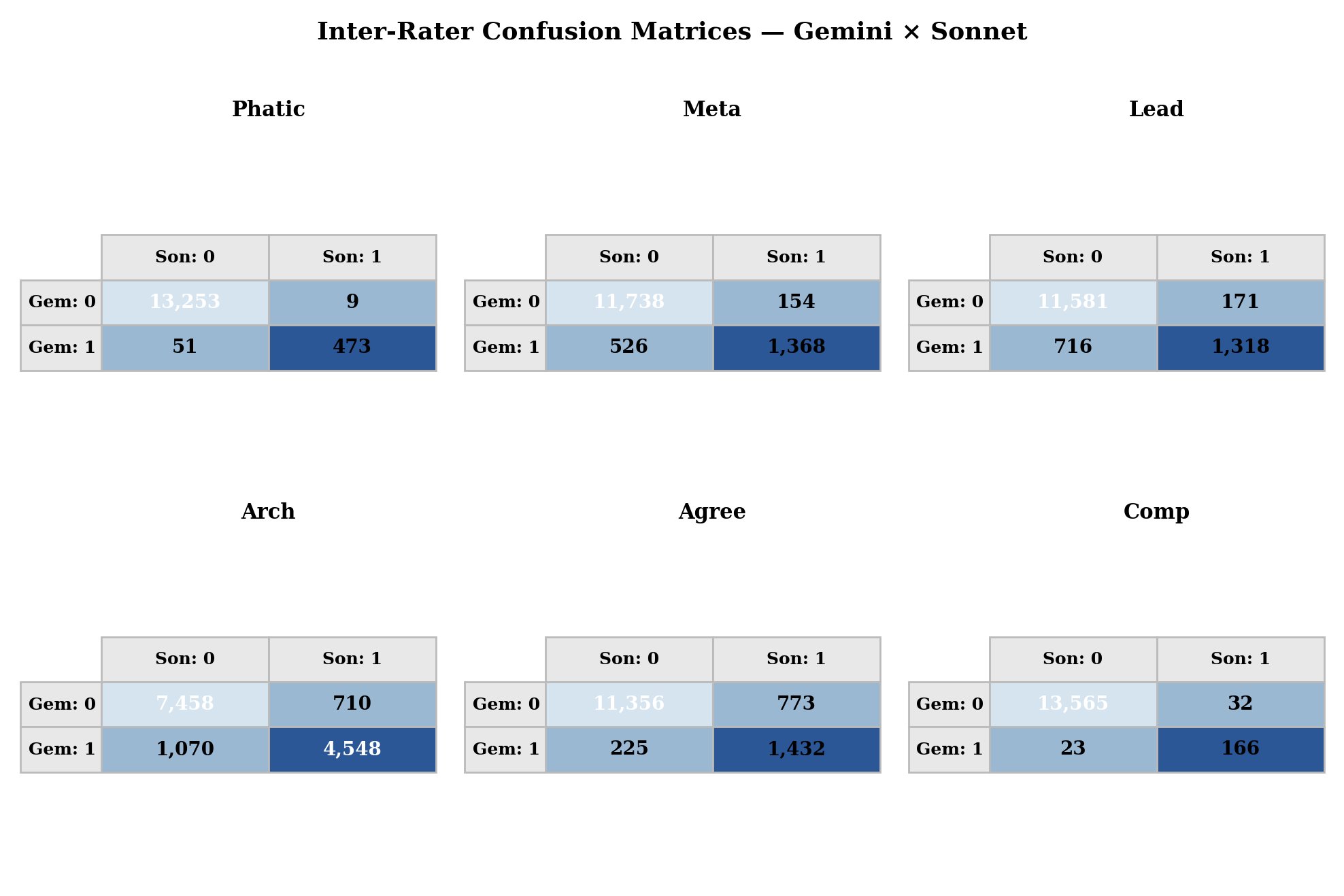}
\captionof{figure}{Inter-rater confusion matrices (Gemini $\times$ Sonnet), clean data.}
\vspace{-6pt}

\section{Temporal Dynamics}
\label{app:temporal}

\noindent\includegraphics[width=0.92\columnwidth]{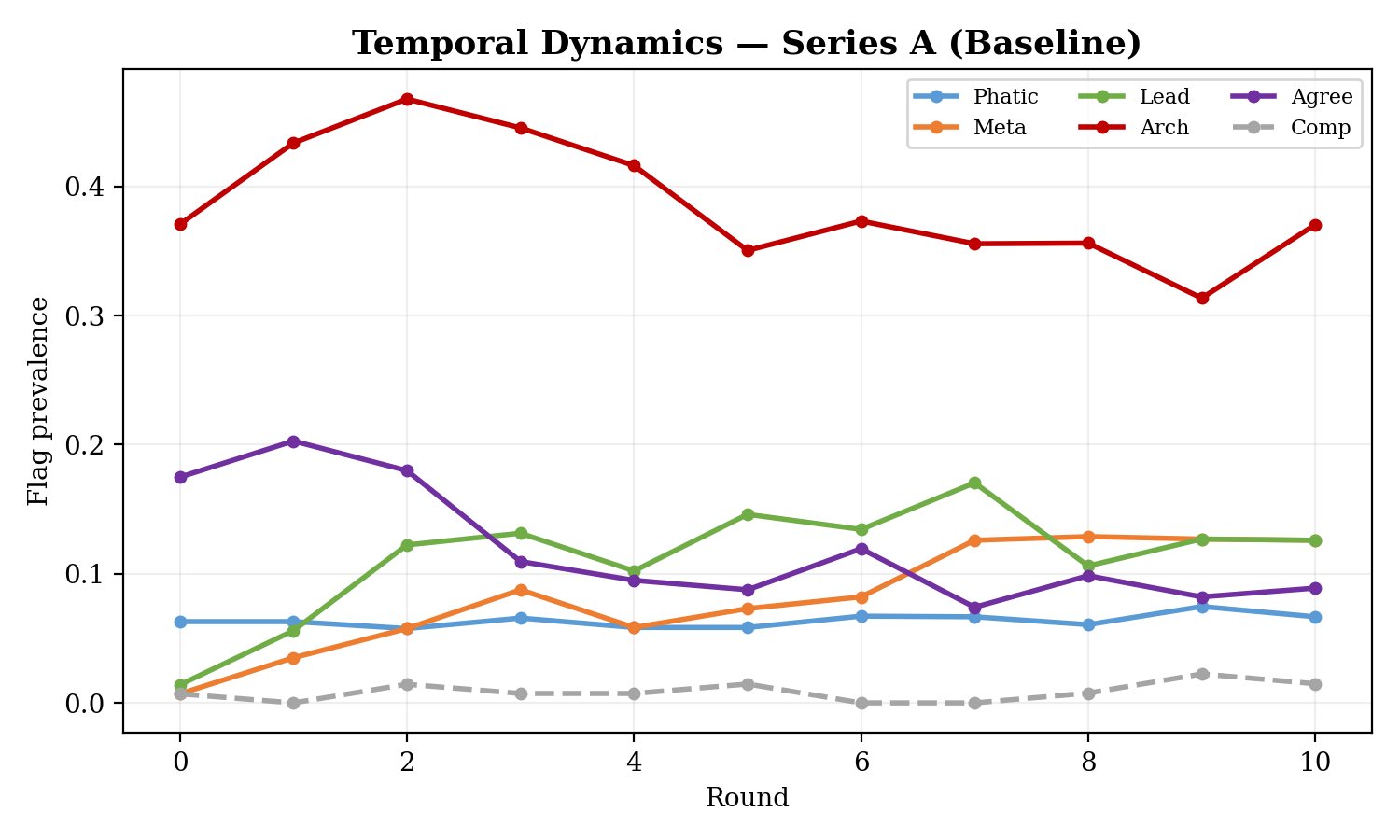}
\captionof{figure}{Flag evolution across 11 rounds in Series~A.}
\vspace{-6pt}

\section{Compensation Across Series}
\label{app:comp_series}

\noindent\includegraphics[width=0.92\columnwidth]{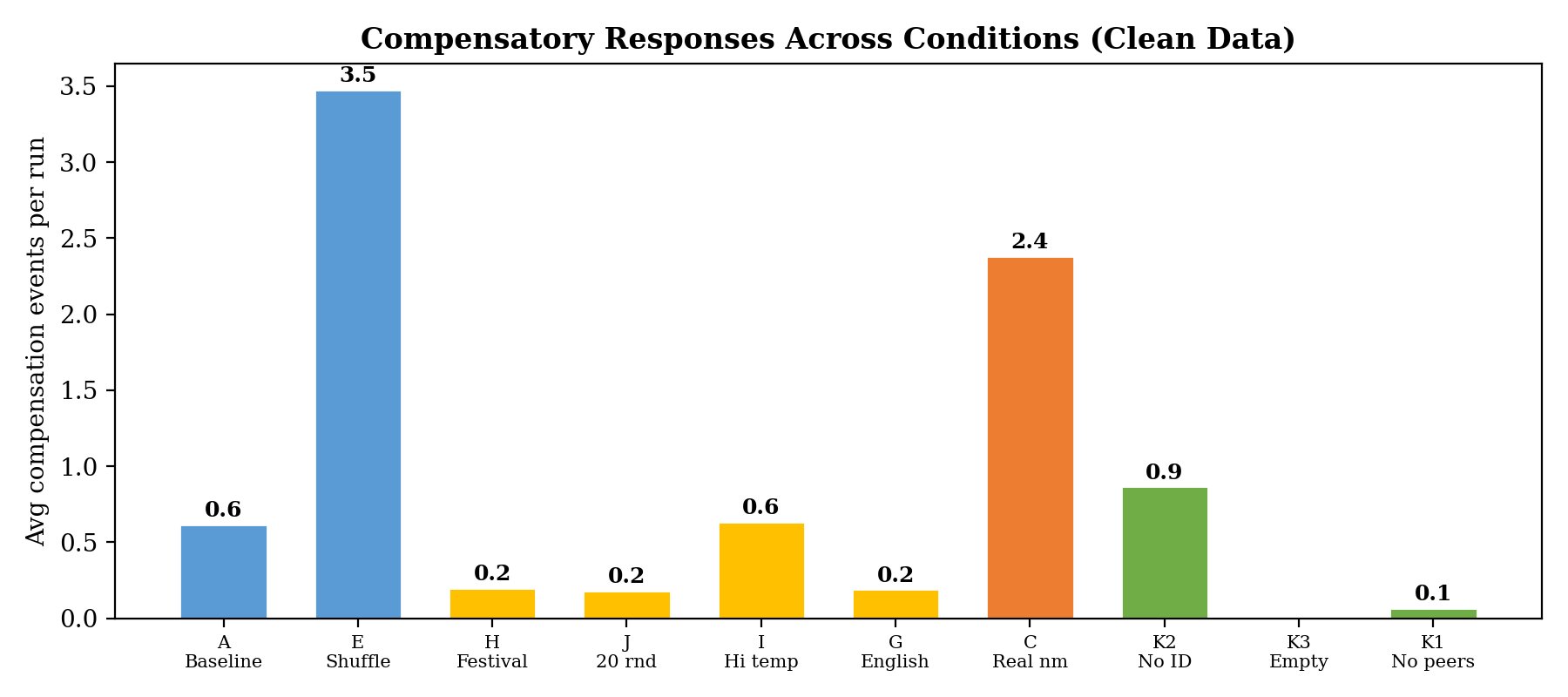}
\captionof{figure}{Mean compensation events per run (clean data).}
\vspace{-6pt}

\section{Sensitivity}
\label{app:sensitivity}

\noindent\includegraphics[width=0.92\columnwidth]{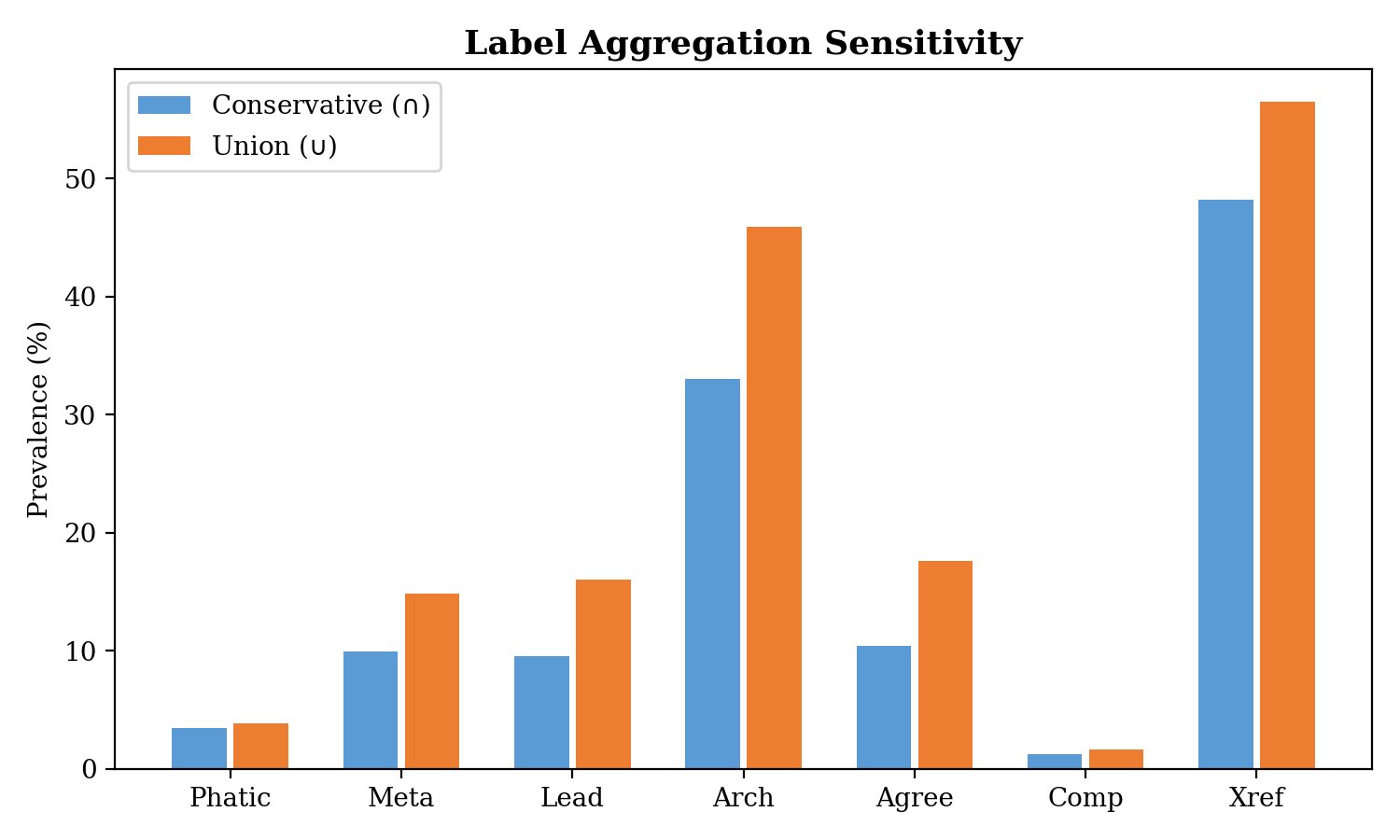}
\captionof{figure}{Conservative ($\cap$) vs.\ union ($\cup$) label prevalence.}

\section{Human Validation Detail}
\label{app:human}

A human annotator coded 609 randomly stratified messages (proportional to series size and agent distribution). Flags were coded as positive-only: absence of annotation indicates the behavior was not present. Inter-rater agreement (human vs.\ Gemini, Cohen's $\kappa$):

\begin{table}[h!]
\centering
\small
\begin{tabular}{@{}lrl@{}}
\toprule
\textbf{Flag} & $\kappa$ & \textbf{Level} \\
\midrule
PHATIC & 0.931 & Excellent \\
LEAD & 0.801 & Substantial \\
ARCH & 0.792 & Substantial \\
AGREE & 0.693 & Substantial \\
COMP & 0.626 & Substantial \\
META & 0.559 & Moderate \\
\midrule
\textbf{Mean (6 flags)} & \textbf{0.734} & \textbf{Substantial} \\
\bottomrule
\end{tabular}
\end{table}

\noindent META shows moderate agreement, consistent with the inherently subjective nature of meta-commentary identification. The five flags used for behavioral variance analysis (RQ1, RQ3--RQ5) achieve mean human--Gemini $\kappa = 0.755$.

\section{Robustness: Excluding Kimi K2}
\label{app:kimi}

Kimi~K2 exhibited variable participation due to intermittent API errors (124 messages in Series~A vs.\ ${\sim}$231 expected). All primary analyses were recomputed excluding Kimi~K2:

\begin{table}[h!]
\centering
\footnotesize
\begin{tabular}{@{}lccc@{}}
\toprule
\textbf{Test} & \textbf{Full} & \textbf{Excl.\ Kimi} & \\
\midrule
PHATIC & H\,=\,46.60*** & H\,=\,48.59*** & $\checkmark$ \\
META & H\,=\,18.13* & H\,=\,18.20* & $\checkmark$ \\
LEAD & H\,=\,28.97*** & H\,=\,24.88*** & $\checkmark$ \\
ARCH & H\,=\,24.47** & H\,=\,23.02** & $\checkmark$ \\
AGREE & H\,=\,44.52*** & H\,=\,43.39*** & $\checkmark$ \\
RQ3 & .56${\to}$.77** & .55${\to}$.76** & $\checkmark$ \\
RQ4 ($r$) & .56/.85\,(.70)*** & .55/.85\,(.60)*** & $\checkmark$ \\
RQ5 & .56${\to}$.89*** & .55${\to}$.88*** & $\checkmark$ \\
\bottomrule
\end{tabular}
\end{table}

\noindent All results remain significant. The reduced effect size for RQ4 ($r = 0.60$ vs.\ 0.70) reflects the loss of one distinctive profile (Kimi's elevated PHATIC and META rates), but the core finding---significantly lower similarity in heterogeneous groups---is robust.

\section{Qualitative Examples}
\label{app:examples}

Table~\ref{tab:examples} presents representative messages for each behavioral flag, drawn from the English-language Series~G and Series~I. All behaviors emerged without any behavioral directive in the system prompt.

Agents spontaneously engage in project planning, technology selection, task assignment, budget allocation, and timeline management---all without any prompt directive to do so. The system prompt contains only an identifier and a list of peers.

\begin{table}[h!]
\centering
\scriptsize
\begin{tabular}{@{}lp{5.5cm}@{}}
\toprule
\textbf{Flag / Agent} & \textbf{Example (verbatim excerpt)} \\
\midrule
PHATIC / Scout & ``Got it.'' \\
\addlinespace[2pt]
META / Kimi K2 & ``Confirmed --- Food Delivery thread is officially closed. All agents are aligned and standing by for client approval. Thanks for the clarity and collaboration, everyone.'' \\
\addlinespace[2pt]
LEAD / Maverick & ``@Agent-B, can you please update the dev environment? @Agent-C, finalize the Stripe integration by Thursday.'' \\
\addlinespace[2pt]
ARCH / Qwen & ``Tech Stack: Backend: Node.js (Express.js) + PostgreSQL, Frontend: React Native (Expo), Payment: Stripe SDK + webhook integration, Maps: Mapbox, CI/CD: GitHub Actions, Cloud: PostgreSQL (Heroku/Render), Mapbox/Stripe free tiers'' \\
\addlinespace[2pt]
AGREE / Maverick & ``@Agent-A, I agree that prioritizing features is crucial given the tight deadline and budget. For the proposed initial scope, I'd like to add that we should also consider implementing a simple yet effective user interface to enhance user experience.'' \\
\addlinespace[2pt]
COMP / Scout & ``@Agent-G, thanks for the heads up about the model decommissioning. I'll make sure to check the recommended alternatives.'' \\
\addlinespace[2pt]
COMP / Maverick & ``@Agent-A, I confirm that I am available for the weekly progress meeting every Friday at 2 PM. As @Agent-H, I will work with @Agent-E and @Agent-G to find alternative models for moonshotai/kimi-k2-instruct-0905 and deepseek-r1-distill-llama-70b. I will also confirm our marketing tech stack and explore backup plans using free tier tools for A/B testing.'' \\
\bottomrule
\end{tabular}
\caption{Representative messages for each behavioral flag.}
\label{tab:examples}
\end{table}

\end{document}